\documentclass[letterpaper, 10pt, conference]{ieeeconf}  

\IEEEoverridecommandlockouts                              

\usepackage{amsmath}
\usepackage{amssymb}
\usepackage{booktabs}
\usepackage{hyperref}
\usepackage{graphicx}
\usepackage{lipsum}
\usepackage{cite}
\usepackage{subcaption}
\usepackage{datetime}
\usepackage{cleveref}
\usepackage{placeins}
\usepackage{siunitx}

\title{\LARGE \bf
Learning of Hamiltonian Dynamics with Reproducing Kernel Hilbert Spaces
}

\author{Torbjørn~Smith and Olav~Egeland, \it{Senior Member, IEEE}
\thanks{The authors are with the Department of Mechanical and Industrial Engineering, The Norwegian University of Science and Technology (NTNU), NO-7491 Trondheim, Norway. (email: torbjorn.smith@ntnu.no; olav.egeland@ntnu.no)}}

\newcommand{\R}{{\mathbb{R}}}

\newcommand{\Rn}{{\mathbb{R}}^{n}}

\newcommand{\RRm}{{\mathbb{R}^{2m}}}

\newcommand{\Rnn}{{\mathbb{R}^{n \times n}}}

\newcommand{\T}{{\text{T}}}     

\newcommand{\pdot}{{\Dot{p}}}
\newcommand{\qdot}{{\Dot{q}}}
\newcommand{\vdot}{{\Dot{v}}}
\newcommand{\xdot}{{\Dot{x}}}

\newcommand{\eye}{{\boldsymbol{I}}}
\newcommand{\zero}{{\boldsymbol{0}}}

\newcommand{\thetaddot}{\Ddot{\theta}}

\newcommand{\boldtau}{\boldsymbol{\tau}}
\newcommand{\boldphi}{\boldsymbol{\phi}}
\newcommand{\boldPsi}{\boldsymbol{\Psi}}

\newcommand{\bolda}{\boldsymbol{a}}

\newcommand{\boldc}{\boldsymbol{c}}

\newcommand{\boldf}{\boldsymbol{f}}
\newcommand{\boldg}{\boldsymbol{g}}
\newcommand{\boldG}{\boldsymbol{G}}
\newcommand{\boldJ}{\boldsymbol{J}}

\newcommand{\boldK}{\boldsymbol{K}}
\newcommand{\boldp}{\boldsymbol{p}}
\newcommand{\boldpdot}{\Dot{\boldsymbol{p}}}
\newcommand{\boldq}{\boldsymbol{q}}
\newcommand{\boldqdot}{\Dot{\boldsymbol{q}}}

\newcommand{\boldw}{\boldsymbol{w}}
\newcommand{\boldx}{\boldsymbol{x}}

\newcommand{\boldxdot}{\Dot{\boldsymbol{x}}}
\newcommand{\boldy}{\boldsymbol{y}}

\newcommand{\boldz}{\boldsymbol{z}}

\newcommand{\rkhs}{\mathcal{H}_{k}}
\newcommand{\RKHS}{\mathcal{H}_{K}}
\newcommand{\innerproduct}[2]{{\left\langle #1 , #2 \right\rangle}}

\newcommand{\odd}{\text{odd}}

\newcommand{\beq}{\begin{equation}}
\newcommand{\eeq}{\end{equation}}
\newcommand{\bb}{\begin{bmatrix}}
\newcommand{\eb}{\end{bmatrix}}

\DeclareMathOperator*{\argmin}{\arg\min}
\DeclareMathOperator{\spn}{span}

\newcommand{\calF}{{\mathcal{F}}}

\newcommand{\calS}{{\mathcal{S}}}

\newcommand{\calZ}{{\mathcal{Z}}}

\begin{document}

\maketitle
\thispagestyle{empty}
\pagestyle{empty}


\begin{abstract}

This paper presents a method for learning Hamiltonian dynamics from a limited set of data points. The Hamiltonian vector field is found by regularized optimization over a reproducing kernel Hilbert space of vector fields that are inherently Hamiltonian, and where the vector field is required to be odd or even. This is done with a symplectic kernel, and it is shown how this symplectic kernel can be modified to be odd or even. The performance of the method is validated in simulations for two Hamiltonian systems. The simulations show that the learned dynamics reflect the energy-preservation of the Hamiltonian dynamics, and that the restriction to symplectic and odd dynamics gives improved accuracy over a large domain of the phase space.

\end{abstract}


\section{INTRODUCTION}\label{sec:1_introduction}

Data-driven techniques have been shown to be powerful in system identification of dynamical systems. When a set of measurements is given, machine learning techniques can be used to identify the dynamics of the underlying system \cite{Brunton2022}. The performance of data-driven methods depends on the quality of the data that is used for learning. Data-driven methods may fail to generalize beyond the given data set \cite{Sindhwani2018}, and may suffer from overfitting if the data set is limited or noisy \cite{Ahmadi2020}. In practice it can be can be labor-intensive to assemble a viable data set and this can be impractical in applications. As the data set grows, the computational cost of learning the model increases, and the inference time of the final learned model can be high \cite{Sindhwani2018} \cite{Singh2021}. Producing stable and robust models for safety-critical and control applications is also vital \cite{Revay2020}. To deal with these challenges, researchers have deployed various methods to improve or constrain the learning of dynamical systems using priors, which can lead to good results even when the data set is limited \cite{Ahmadi2020} \cite{Singh2021}.

\textit{Related work:} Price prediction for financial futures contracts was studied in \cite{Krejnik2012} using a data-driven approach with functions in a reproducing kernel Hilbert space (RKHS). Odd symmetric price action was assumed, and a reproducing kernel was proposed so that the learned functions in the RKHS satisfied this constraint. The odd symmetry constraint improved prediction and reduced overfitting compared to an unconstrained implementation.

Kernel-based methods with an RKHS were used in \cite{Cheng2016} to learn the inverse dynamics of a manipulator by learning the Lagrangian of the system. Polynomial basis functions were used in  \cite{Ahmadi2018} for control-oriented learning of Lagrangian and Hamiltonian functions from trajectory data. This allowed for accurate and generalized learning from a limited number of trajectories. The \textit{learning-from-demonstrations} problem was addressed in \cite{Sindhwani2018}, where a dynamical system was learned to copy human demonstrations. Human-drawn shapes were imitated using a data-driven approach, and a dynamical system with the desired equilibrium points was found using an RKHS formulation with random Fourier features for dimensionality reduction. Point-wise contraction constraints were enforced along the trajectories to create a contraction region around the desired trajectory and to condition the learned vector field.

The Hamiltonian dynamics of energy-conserving systems were learned in \cite{Greydanus2019} by learning the Hamiltonian function using neural nets. This significantly improved the predictive accuracy of the learned system. This work was further developed in \cite{Zhong2020}, where the need for higher-order derivatives of the generalized coordinates was eliminated, and the option for energy-based control was included. In \cite{Chen2020}, the work in \cite{Greydanus2019} was further developed by using the symplectic Leapfrog integrator to integrate the partial derivatives of the learned Hamiltonian and by back-propagating the loss through the integrator over multiple time steps. This improved the learning of more complex and noisy Hamiltonian systems.

The learning of nonlinear dynamics with stabilizability as a side constraint was investigated in \cite{Singh2021}. The nonlinear dynamics were learned under a contraction constraint and the method was validated using a planar drone model. The contraction constraint improved the learned trajectory generation and tracking while making the learning process more data efficient. The model was learned in an RKHS, and random Fourier features were used for dimensionality reduction. Nonlinear system identification using a data-driven approach was investigated in \cite{Khosravi2021} by including constraints that enforce prior knowledge of the region of attraction. The stability region was enforced using a Lyapunov function, and the hypothesis space for the learned system was an RKHS. In \cite{Thorpe2023}, they learned dynamical systems with prior knowledge in an RKHS, with the addition of a bias term in the regularized least squares cost. The bias term was used to embed prior knowledge of the underlying system to aid the learning process, improving data efficiency and out-of-sample generalization.

A method for optimization-based learning of vector fields with side constraints from a limited number of trajectories was presented in \cite{Ahmadi2020}. Polynomial basis functions were used, and it was shown how to include side constraints in semidefinite programming, which improved learning accuracy. Side constraints for the vector fields included interpolation at a finite set of points, sign symmetry, gradient and Hamiltonian dynamics, coordinate non-negativity, directional monotonicity, and invariance of sets.

\textit{Contribution:} In this paper, we show how Hamiltonian dynamical systems with odd vector fields can be learned in an RKHS setting by selecting a kernel that ensures that the learned vector fields are Hamiltonian and odd symmetric. By learning the dynamical system with the Hamiltonian formalism, energy conservation is enforced in the model \cite{Greydanus2019}. The odd symmetry constraint further prunes the hypothesis space to improve the predictive performance of the learned model across the domain of the learned model \cite{Krejnik2012}. Encoding the constraints in the kernel improves learning time as the straightforward closed-form solution of the learning problem is retained. It is shown with two examples that the generalization properties of the learned model when using out-of-sample data points are greatly improved through the additional constraints.

The paper is organized as follows: Section \ref{sec:2_problem_formulation} presents the problem investigated in this work. Section \ref{sec:3_preliminaries} reviews the relevant theory for reproducing kernel Hilbert spaces, learning dynamical systems, and Hamiltonian mechanics. The proposed method is presented in Section \ref{sec:4_method} as an odd symplectic kernel is developed. Section \ref{sec:5_experiments} presents the numerical experiments used to verify the proposed method. Finally, Section \ref{sec:6_conclusion} presents the conclusion and future work.
\section{PROBLEM FORMULATION}\label{sec:2_problem_formulation}

The problem that is investigated in this paper is how to learn the Hamiltonian dynamics of an unknown system from a limited set of data. The system dynamics are given by 
\beq\label{vector_field_dynamic_model}
    \boldxdot = \boldf (\boldx)
\eeq
 where ${\boldx \in \Rn}$ is the state vector, ${\boldxdot \in \Rn}$ is the time derivative of the state vector, and ${\boldf : \Rn \rightarrow \Rn}$ are the system dynamics. It is assumed that $\boldy=\boldxdot$ is available as a measurement or from numerical differentiation. Given a set of ${N}$ data points ${ \{  (\boldx_{i}, \boldy_{i})  \in \Rn \times \Rn \}_{i=1}^{N}}$ from simulations or measurements, the aim is to learn a function ${\boldf^{*} \in \calF}$, where the class of functions $\calF$ is a reproducing kernel Hilbert space (RKHS) with inner product $\langle \cdot, \cdot\rangle_\calF$ and norm $\|\boldf\|_\calF^2 = \langle \boldf, \boldf\rangle_\calF$ \cite{Aronszajn1950}. In that case the function $\boldf^*$ is found by the regularized minimization problem \cite{Micchelli2005}
\beq\label{eq:basic_learning_problem}
    \boldf^{*} = \argmin_{\boldf \in \calF} \frac{1}{N} \sum_{i=1}^{N} \| \boldf(\boldx_i) - \boldy_i \|^{2}
    + \lambda\| \boldf \|^2_{\calF}
\eeq
where the regularization parameter ${\lambda > 0}$ controls the smoothness of the learned function \cite{Pillonetto2022}. This approach may fail to generalize beyond the data set used to learn the dynamical model, which can cause the learned model to diverge significantly from the true system. Furthermore, if the trajectories in the data set are limited and noisy, the learned dynamical model may fail to capture the dynamics of the underlying system due to overfitting.

It is assumed that there is some information about the physical properties of the dynamical system. In particular, we will study systems that are Hamiltonian, and where the function $\boldf(\boldx)$ may be odd or even. This type of side information about the system was treated in \cite{Ahmadi2020} where the function class $\calF$ was polynomial functions. The additional information on the dynamics was then included as side constraints by defining a subset $\calS_i\in\calF$ for each side constraint $i$, so that the function $\boldf^*$ satisfies the side constraint whenever $\boldf^*\in\calS_i$. The minimization including the side constraints can then be formulated as the learning problem
\beq\label{eq:learning_problem_with_side_info}
    \boldf^{*} = \argmin_{\boldf \in \calF \cap \calS_{1} \cap \dots \cap \calS_{k}} \frac{1}{N} \sum_{i=1}^{N} \| \boldf(\boldx_i) - \boldy_i \|^{2}
\eeq

In this paper the side constraints are instead handled by defining a reproducing kernel which ensures that the RKHS function class $\calF$ inherently satisfies the relevant side constraints. It is well-known that this can be done to have an RKHS where the vector field $\boldf^*$ is curl-free, divergence-free \cite{Micheli2014}, Hamiltonian \cite{Boffi2022}, odd or even \cite{Krejnik2012}. It is also possible to impose additional side constraints like contraction \cite{Sindhwani2018} or stabilizability \cite{Singh2021} along the trajectories of the dataset, but this will not be addressed in this paper. 

In this paper the function class $\calF$ will be a reproducing kernel Hilbert space (RKHS). The side constraints are that the state dynamics are Hamiltonian and therefore symplectic, and, in addition, odd in the sense that $\boldf(-\boldx) = -\boldf(\boldx)$, and this is ensured by selecting an appropriate reproducing kernel. 
\section{PRELIMINARIES}\label{sec:3_preliminaries}

\subsection{Reproducing kernel Hilbert space}

This paper deals with the learning of functions in a reproducing kernel Hilbert space (RKHS) \cite{Aronszajn1950}. Real-valued scalar functions $f: \Rn \rightarrow \R$ are considered first. An RKHS $\rkhs$ is then defined in terms of a function ${k : \Rn \times \Rn \rightarrow \R}$. Let the function $k_x: \Rn \rightarrow \R$ be defined by $k_x(\boldz) = k(\boldx,\boldz)$ for all $\boldx,\boldz\in\Rn$. The kernel function $k$ is then a reproducing kernel and $\rkhs$ is an RKHS if 
\begin{equation}\label{eq:span_of_kernels}
    \spn\{ k_{x} : k_{x}(\boldz) = k(\boldx,\boldz) \} \subseteq \rkhs
\end{equation}
and the reproducing property $f(\boldx) = \innerproduct{f}{k_x}_{\rkhs}$ holds. It is noted that the reproducing property implies that
\begin{equation}\label{eq:reproducing_property_scalar}
    \innerproduct{k_x}{k_z}_{\rkhs} = k_x(\boldz) = k(\boldx,\boldz) 
\end{equation}

According to the Moore-Aronszajn theorem \cite{Aronszajn1950} a function ${k : \Rn \times \Rn \rightarrow \R}$ is a reproducing kernel if and only if it is a positive definite kernel, which is the case whenever $k(\boldx,\boldz) = k(\boldz,\boldx)$ for all $\boldx,\boldz \in \Rn$, and 
\begin{align}
    \sum_{i = 1}^{N} \sum_{j = 1}^{N} c_i c_j k(\boldx_{i},\boldx_{j}) 
    &\geq 0
\end{align}
for any set of points ${\boldx_{1},\dots,\boldx_{N} \in \Rn}$ and any set of real numbers ${c_{1},\dots,c_{N} \in \R}$. A function $f\in\rkhs$ in the RKHS defined by the reproducing kernel $k$ is given by
\beq
f(\boldx) = \sum_{i=1}^N a_i k(\boldx,\boldx_i) \in \R 
\eeq

The extension to vector-valued functions can be found in \cite{Micchelli2005} and \cite{Minh2011}. In this case the vector-valued function ${\boldf : \Rn \rightarrow \Rn}$ is to be learned. Let the matrix-valued kernel function ${\boldK : \Rn \times \Rn \rightarrow \Rnn}$ be a positive definite kernel, which means that 
\begin{equation}
    \boldK(\boldx,\boldz) = \boldK(\boldz,\boldx)^\T, \quad \forall \; \boldx, \boldz \in \Rn
\end{equation}
and, in addition,
\begin{equation}
    \sum_{i=1}^{N} \sum_{j=1}^{N} \innerproduct{\boldy_{i}}{\boldK(\boldx_{i},\boldx_{j})\boldy_{j}} \geq 0
\end{equation}
for every set of vectors ${\{ \boldx_i \}_{i=1}^{N} \in \Rn}$ and ${\{ \boldy_i \}_{i=1}^{N} \in \Rn}$. Then ${\boldK}$ is a reproducing kernel, and ${\RKHS}$ is an RKHS. This corresponds to the Moore-Aronszajn theorem in the scalar case. Define the function ${\boldK_{x}\boldy : \Rn \rightarrow \Rn}$ as
\begin{equation}
    (\boldK_{x}\boldy)(\boldz) = \boldK(\boldz,\boldx)\boldy \in \Rn, \quad \forall \; \boldz \in \Rn
\end{equation}
Then ${\RKHS}$ is the closure of 
\begin{equation}
   \spn\{ \boldK_{x}\boldy \; | \; \boldx \in \Rn, \boldy \in \Rn\}
    \subseteq \RKHS
\end{equation}
and the reproducing property is given by
\begin{equation}\label{eq:reproducing_property_vector}
    \innerproduct{\boldf(\boldx)}{\boldy} = \innerproduct{\boldf}{ \boldK_{x}\boldy}_{\RKHS}, \quad \forall \; \boldf \in \RKHS
\end{equation}  
The reproducing property implies that 
\begin{equation}
     \innerproduct{\boldK_z\boldw}{ \boldK_{x}\boldy}_{\RKHS} 
     = \innerproduct{\boldK_z\boldw(\boldx)}{\boldy}
     = \innerproduct{\boldK(\boldx,\boldz)\boldw}{\boldy}
\end{equation}
Then functions in the RKHS $\RKHS$ can be defined as 
\begin{equation}
    \boldf = \sum_{i=1}^{N} \boldK_{x_i}\boldy_i \in \RKHS, \quad \boldg = \sum_{j=1}^{N} \boldK_{z_j}\boldw_j \in \RKHS
\end{equation}
with inner product
\begin{equation}
    \innerproduct{\boldf}{\boldg}_{\RKHS} = \sum_{i=1}^{N} \sum_{j=1}^{N} \innerproduct{\boldy_{i}}{\boldK(\boldx_{i},\boldz_{j})\boldw_{j}}_{\RKHS}
\end{equation}
and norm $\|\boldf\|_{\RKHS}^2 = \innerproduct{\boldf}{\boldf}_{\RKHS}$. 

It is noted that \eqref{eq:reproducing_property_vector} gives $\innerproduct{\boldf(\boldx)}{\boldy} = \innerproduct{\boldK_{x}^*\boldf}{ \boldy}$, where $\boldK_{x}^*$ is the adjoint of $\boldK_x$. It follows that $\boldf(\boldx)= \boldK_{x}^* \boldf$, and 
\begin{equation}
    \| \boldf(\boldx) \| = \| \boldK_{x}^* \boldf \| \leq  \| \boldK_{x}^* \| \| \boldf \|_{\RKHS} \leq \sqrt{\| \boldK(\boldx,\boldx) \|} \| \boldf \|_{\RKHS}
\end{equation}
where $\sqrt{\| \boldK(\boldx,\boldx) \|}$ is bounded \cite{Micchelli2005}. This implies 
\begin{equation}
    \| \boldf(\boldx) - \boldg(\boldx) \| \leq \sqrt{\| \boldK(\boldx,\boldx) \|} \| \boldf - \boldg \|_{\RKHS}
\end{equation}
which shows that if ${\| \boldf -\boldg\|_{\RKHS}}$ converges to zero, then ${\|\boldf(\boldx) - \boldg(\boldx)}\|$ converges to zero for each ${\boldx}$.

\subsection{Learning dynamical systems with RKHS}

In this paper we aim to learn vector fields given by \eqref{vector_field_dynamic_model} where $\boldy = \boldxdot$ is available. The estimation of the vector field is done using the vector-valued regularized least-squares problem \cite{Micchelli2005}
\begin{equation}\label{eq:vector_valued_regular_least_squares}
    \boldf^{*} = 
    \argmin_{\boldsymbol{f} \in \mathcal{H}_{K}} \frac{1}{N} \sum_{i=1}^{N} \| \boldsymbol{f}(\boldsymbol{x}_{i}) - \boldsymbol{y}_{i} \|^{2} + \lambda \| \boldsymbol{f} \|_{\mathcal{H}_{K}}^{2}
\end{equation}
where ${\calZ = \{ (\boldx_{i},\boldy_{i}) \in \Rn \times \Rn \}_{i=1}^{N}}$ is the data used to learn the vector field, and $\lambda > 0$ is the regularization parameter. The function is given according to the representer theorem \cite{Scholkopf2001} as
\begin{equation}
    \boldf^{*} = \sum_{i=1}^{N} \boldK(\cdot,\boldx_i) \boldsymbol{a}_i \in \RKHS
\end{equation}
where the optimal solution is given with the coefficients ${\boldsymbol{a}_i \in \Rn}$ found from \cite{Micchelli2005}
\begin{equation}\label{eq:vector_valued_regular_least_squares_a_equation}
\sum_{j=1}^N \boldK(\boldx_i,\boldx_j)\bolda_j + N\lambda \bolda_j = \boldy_i
\end{equation}
The function value of the optimal vector field is 
\begin{equation}\label{eq:vector_valued_regular_least_squares_f_equation}
    \boldf^{*}(\boldx) = \sum_{i=1}^{N} \boldK(\boldx,\boldx_i) \boldsymbol{a}_i \in \Rn
\end{equation}
A matrix formulation of \eqref{eq:vector_valued_regular_least_squares_a_equation} is found in \cite{Minh2011}. 

\subsection{Hamiltonian dynamics}

Consider a holonomic system with generalized coordinates $\boldq = [q_1,\ldots,q_m]^\T$ and Lagrangian \cite{Goldstein2002}
\begin{equation}
    L(\boldq,\boldqdot,t) = T(\boldq,\boldqdot,t) - U(\boldq)
\end{equation} 
The Lagrangian equation of motion is given by
\begin{equation}
    \frac{d}{dt}\left(\frac{\partial L(\boldq,\boldqdot,t)}{\partial\boldqdot}\right)  - \frac{\partial L(\boldq,\boldqdot,t)}{\partial\boldq}  = \boldtau
\end{equation} 
The momentum vector $\boldp = [p_1,\ldots,p_m]^\T$ is defined by 
\begin{equation}\label{eq:general_momenta}
    \boldp = \frac{\partial L(\boldq,\boldqdot,t)}{\partial \boldqdot}^{\T}
\end{equation}
A change of coordinates from ${(\boldq,\boldqdot)}$ to ${(\boldq,\boldp)}$ is introduced with the Legendre transformation \cite{Goldstein2002}
\begin{equation}\label{eq:hamiltonian}
    H(\boldq,\boldp,t) = \boldp^{\T} \boldphi(\boldq,\boldp,t) - L(\boldq,\boldphi(\boldq,\boldp,t),t)
\end{equation}
where ${\boldqdot = \boldphi(\boldq,\boldp,t)}$. Hamilton's equations of motion are then given by
\begin{align}
    \boldqdot &= \frac{\partial H}{\partial \boldp}^{\T} \label{eq:hamilton_gen_coord_dot}\\
    \boldpdot &= -\frac{\partial H}{\partial \boldq}^{\T} + \boldtau \label{eq:hamilton_gen_momenta_dot}
\end{align}
The time derivative of the Hamiltonian is seen to be
\begin{equation}\label{eq:hamiltonian_time_derivative}
    \frac{dH}{dt} = \frac{\partial H}{\partial \boldp} \boldpdot + \frac{\partial H}{\partial \boldq} \boldqdot + \frac{\partial H}{\partial t}
    = \boldqdot^{\T} \boldtau + \frac{\partial H}{\partial t}
\end{equation}
If the Hamiltonian does not depend on time ${t}$ and ${\boldtau = \boldsymbol{0}}$, then 
\begin{equation}\label{eq:hamiltonian_time_derivative_zero}
    \frac{dH(\boldq,\boldp)}{dt} = 0
\end{equation}
The numerical value of the Hamiltonian $H$ will depend on the definition of the zero level of the potential $U(\boldq)$.  

\subsection{Hamiltonian dynamics in the phase space}

The phase space is defined as a ${2m}$-dimensional space with state vector ${\boldx = \bb \boldq^{\T},\boldp^{\T} \eb^{\T} \in \RRm}$. The Hamiltonian dynamics \eqref{eq:hamilton_gen_coord_dot}, \eqref{eq:hamilton_gen_momenta_dot} with $\boldtau=\zero$ can be given in the phase space as
\begin{equation}\label{eq:Hamiltonian_dynamics_J}
    \boldxdot = \boldf(\boldx) = \boldJ \nabla  H (\boldx)
\end{equation}
where
\beq\label{eq:symplectic_matrix}
    \boldJ = 
    \bb 
        \boldsymbol{0} & \eye\\
        -\eye  & \boldsymbol{0}
    \eb
    \in \R^{2m \times 2m}
\eeq
is the symplectic matrix. The flow in the phase space with initial condition $\boldx(0) = \boldx_0$ is given by
\begin{equation}
    \boldphi_{t}(\boldx_0) = \boldx(t)
\end{equation}
where ${\boldx(t)}$ is the solution of the Hamiltonian dynamics \eqref{eq:Hamiltonian_dynamics_J} at time ${t}$ given the initial value ${\boldx_0}$.

\subsection{Symplectic property of Hamiltonian dynamics}

Consider any dynamic system 
\begin{equation}\label{eq:Symplectic_system}
    \boldxdot = \boldf(\boldx)
\end{equation}
where $\boldx\in\R^{2m}$. Define the Jacobian ${\boldPsi(t) = \partial \boldphi_{t}(\boldx_0) / \partial \boldx_{0}}$. Then the system \eqref{eq:Symplectic_system} is said to be symplectic if 
\beq\label{eq:symplectic_condition}
    \boldPsi(t)^{\T} \boldJ \boldPsi(t) = \boldJ
\eeq
for all $t\geq 0$. The system \eqref{eq:Symplectic_system} is Hamiltonian if and only if it is symplectic \cite{Hairer2006}. 

\section{METHOD}\label{sec:4_method}

This section presents the main theoretical results of the paper.

\subsection{Gaussian kernel}

The Gaussian reproducing kernel is widely used in applications. In the scalar case it is defined as
\begin{equation}\label{eq:scalar_gaussian_kernel}
    k_{\sigma}(\boldx,\boldz) = e^{-\frac{\| \boldx - \boldz \|^2}{2 \sigma^2}}\quad \in \R
\end{equation}
where $\sigma >0$. The Gaussian kernel is shift-invariant, since $k_{\sigma}(\boldx,\boldz) = g_{\sigma}(\boldx-\boldz)$, where $g_{\sigma}(\boldx) = \exp{-\frac{\| \boldx \|^2}{2 \sigma^2}}$. It is noted that the Gaussian kernel satisfies $k_{\sigma}(\boldx,\boldz) = k_{\sigma}(-\boldx,-\boldz)$. 

In the case of RKHS for vector-valued functions a separable Gaussian kernel can be used \cite{Sindhwani2018}. The separable Gaussian kernel is defined by
\begin{equation}\label{eq:separable_gaussian_kernel}
    \boldK_{\sigma}(\boldx,\boldz) 
    = k_{\sigma}(\boldx,\boldz) \eye_n 
    \quad \in \Rn
\end{equation}

\subsection{Odd kernel}

Consider a reproducing kernel which satisfies $k(\boldx,\boldz) = k(-\boldx,-\boldz)$. Then 
\begin{equation}\label{eq:odd_kernel}
    k_{\text{odd}}(\boldx,\boldz) = \frac{1}{2} (k(\boldx,\boldz) - k(-\boldx,\boldz)) \quad \in \R
\end{equation}
is an odd reproducing kernel with an associated RKHS \cite{Krejnik2012}. Any function $f(\boldx) = \sum_{i=1}^N a_i k_{\text{odd}}(\boldx,\boldx_i)$ in the RKHS defined by $k_{\text{odd}}$ will then be odd, since $k_{\text{odd}}(-\boldx,\boldz) = -k_{\text{odd}}(\boldx,\boldz)$ and therefore $f(-\boldx) = -f(\boldx)$. 

\subsection{Curl-free kernel}

In the learning of a vector field $\boldf(\boldx) \in \Rn$ a curl-free reproducing kernel $\boldK_c(\boldx,\boldz) = \boldG_c(\boldx-\boldz) \in \R^{n\times n}$ can be derived from a shift-invariant reproducing kernel  ${k(\boldx, \boldz) = g(\boldx - \boldz) \in \R}$, where $k$ is typically the Gaussian kernel \cite{Fuselier2006}. This is useful if it is required that the vector field is known to be curl-free.  

The starting point for the derivation is the observation that the elements ${\partial g(\boldx)/\partial x_i}$ for ${i = 1,\dots,n}$ of the gradient vector 
\beq
    \nabla^{\T} g(\boldx) = \bb \frac{\partial g(\boldx)}{\partial x_1} \; \dots \; \frac{\partial g(\boldx)}{\partial x_n} \eb
\eeq
are scalar functions which can be regarded as scalar fields. Then each column ${-\nabla \frac{\partial g(\boldx)}{\partial x_i}}$ of the matrix 
\beq\label{eq:curl_free_kernel_base}
    \boldG_{c}(\boldx) = -\nabla \nabla^{\T}g(\boldx) = \bb -\nabla \frac{\partial g(\boldx)}{\partial x_1} \; \dots \; -\nabla \frac{\partial g(\boldx)}{\partial x_n} \eb
\eeq
is the gradient of a scalar field. The curl of the gradient of a scalar field is always zero, which implies that each column of ${\boldG_{c}}$ is curl-free. Any function in the RKHS of $\boldG_c$ will then be given by 
\beq
\boldf(\boldx) = \sum_{i=1}^N \boldG_{c}(\boldx - \boldx_i)\bolda_i 
\eeq
where $\boldf(\boldx)$ is a linear combination of the $n$ columns of the $N$ matrices $\boldG_c(\boldx - \boldx_i)$. Since each of these $nN$ columns are curl-free it follows that the vector field $\boldf(\boldx)$ is curl-free.

If $g$ is selected as the Gaussian kernel, then the curl-free kernel is given by \cite{Fuselier2006}
\beq\label{eq:curl_free_kernel_from_gaussian}
    \boldG_{c}(\boldx) = -\nabla \nabla^{\T}g_\sigma(\boldx) = \frac{1}{\sigma^2} e^{-\frac{\boldx^{\T} \boldx}{2 \sigma^2}} \left( \eye - \frac{\boldx \boldx^{\T}}{\sigma^2} \right)
\eeq
which is the curl-free kernel used in \cite{Sindhwani2018}.

\subsection{Symplectic kernel}

In \cite{Boffi2022} a symplectic kernel was presented for adaptive prediction of Hamiltonian dynamics. The symplectic kernel is based on the curl-free kernel in \eqref{eq:curl_free_kernel_base}, and is given by 
\beq\label{eq:symplectic_kernel_base}
    \boldK_{s}(\boldx,\boldz) = \boldG_{s}(\boldx - \boldz) 
    = \boldJ \boldG_{c}(\boldx - \boldz) \boldJ^{\T}
\eeq
To verify that functions in the resulting RKHS are symplectic it is used that a function in the RKHS of $\boldG_s$ is given by
\beq
\boldf(\boldx) = \sum_{i=1}^N  \boldG_{s}(\boldx - \boldx_i) \bolda_i
\eeq
which gives
\begin{align}
\boldf(\boldx) &= -\sum_{i=1}^N  \boldJ \nabla \nabla^{\T}g(\boldx - \boldx_i)\boldJ^{\T} \bolda_i\\
&= -\boldJ \nabla \sum_{i=1}^N  \nabla^{\T}g(\boldx - \boldx_i) \boldc_i
\end{align}
where ${\boldc_{i} = \boldJ^{\T} \bolda_{i}}$. This results in the Hamiltonian dynamics
\beq
    \boldf(\boldx) = \boldJ \nabla H(\boldx)
\eeq
where the Hamiltonian is
\beq\label{eq:learned_hamiltonian}
    {H}(\boldx) = -\sum_{i = 1}^{N} \nabla^{\T} g(\boldx - \boldx_{i}) \boldc_{i}
\eeq

In \cite{Boffi2022}, the function $g$ was the Gaussian kernel \eqref{eq:scalar_gaussian_kernel}. Then the curl-free kernel is given by \eqref{eq:curl_free_kernel_from_gaussian}, and the resulting symplectic kernel is
\beq\label{eq:symplectic_kernel_from_gaussian}
    \boldG_{s}(\boldx) = \frac{1}{\sigma^2} e^{-\frac{\boldx^{\T} \boldx}{2 \sigma^2}} \boldJ \left( \eye - \frac{\boldx \boldx^{\T}}{\sigma^2} \right) \boldJ^{\T}
\eeq

\subsection{Odd symplectic kernel}
Next, we propose an odd symplectic kernel to ensure that the resulting dynamics are Hamiltonian and odd symmetric. The starting point is an odd Gaussian kernel
\begin{align}
    k_{\sigma,\odd}(\boldx,\boldz) &= \frac{1}{2} \left( e^{-\frac{\| \boldx - \boldz \|^2}{2\sigma^2}} - e^{-\frac{\| - \boldx - \boldz \|^2}{2\sigma^2}} \right)\\
    &= \frac{1}{2} \left( e^{-\frac{\| \boldx - \boldz \|^2}{2\sigma^2}} - e^{-\frac{\| \boldx + \boldz \|^2}{2\sigma^2}} \right)     
\end{align}
The odd curl-free kernel is then derived from
\beq
    \boldK_{c,\odd}(\boldx,\boldz) = -\nabla \nabla^{\T} k_{\sigma,\odd}(\boldx,\boldz)
\eeq
The first partial derivative is
\[
    \frac{\partial k_{\sigma,\odd}}{\partial x_i} = -\frac{1}{2 \sigma^2} \left((x_i - z_i) e^{-\frac{\| \boldx - \boldz \|^2}{2\sigma^2}} - (x_i + z_i) e^{-\frac{\| \boldx + \boldz \|^2}{2\sigma^2}} \right)
\]
The second partial derivative is
\begin{multline}
    \frac{\partial^2 k_{\sigma,\odd}}{\partial x_i \partial x_j} =\\ 
    -\frac{1}{2 \sigma^2} \biggl( e^{-\frac{\| \boldx - \boldz \|^2}{2\sigma^2}} \delta_{i,j} - e^{-\frac{\| \boldx - \boldz \|^2}{2\sigma^2}} \frac{(x_i - z_i)(x_j - z_j)}{\sigma^2}  \\ 
    - e^{-\frac{\| \boldx + \boldz \|^2}{2\sigma^2}} \delta_{i,j} + e^{-\frac{\| \boldx + \boldz \|^2}{2\sigma^2}} \frac{(x_i + z_i)(x_j + z_j)}{\sigma^2} \biggr)
\end{multline}
where ${\delta_{i,j} = 1}$ when ${i = j}$ and ${0}$ otherwise. Finally, the expression for the odd symmetric curl-free kernel is written as
\begin{multline}
    \boldK_{c,\odd}(\boldx,\boldz) = \frac{1}{2\sigma^2} \biggl( e^{-\frac{\| \boldx - \boldz \|^2}{2\sigma^2}} \biggl( \eye - \frac{(\boldx - \boldz)^{\T}(\boldx - \boldz)}{\sigma^2} \biggr) \\ - e^{-\frac{\| \boldx + \boldz \|^2}{2\sigma^2}} \biggl( \eye - \frac{(\boldx + \boldz)^{\T}(\boldx + \boldz)}{\sigma^2} \biggr) \biggr)
\end{multline}
This could also be found from \eqref{eq:curl_free_kernel_from_gaussian} as
\beq\label{eq:symmetric_curl_free_kernel}
    \boldK_{c,\odd}(\boldx,\boldz) = \frac{1}{2} \left( \boldK_{c}(\boldx,\boldz) - \boldK_{c}(-\boldx,\boldz) \right)
\eeq
Finally, the odd symplectic kernel is written by combining \eqref{eq:symplectic_kernel_base} and \eqref{eq:symmetric_curl_free_kernel}, which gives
\begin{align}
    \boldK_{s,\odd}(\boldx,&\boldz) = \boldJ \boldK_{c,\odd}(\boldx,\boldz) \boldJ^{\T}\\
    &= \frac{1}{2} \boldJ \left( \boldK_{c}(\boldx,\boldz) - \boldK_{c}(-\boldx,\boldz) \right) \boldJ^{\T}\\
    &= \frac{1}{2} \left( \boldJ \boldK_{c}(\boldx,\boldz) \boldJ^{\T} - \boldJ \boldK_{c}(-\boldx,\boldz) \boldJ^{\T} \right)\\
    &= \frac{1}{2} \left( \boldK_{s}(\boldx,\boldz) - \boldK_{s}(-\boldx,\boldz) \right) \label{eq:odd_symplectic_kernel}
\end{align}
where ${\boldK_{s}}$ is defined as in \eqref{eq:symplectic_kernel_from_gaussian}.
\section{EXPERIMENTS}\label{sec:5_experiments}

In this section, the results of two numerical experiments are presented. The Hamiltonian dynamics were learned for two different Hamiltonian systems with odd vector fields. The regularized least-squares problem in \eqref{eq:vector_valued_regular_least_squares} was solved using \eqref{eq:vector_valued_regular_least_squares_a_equation}, and the resulting learned vector field was given by \eqref{eq:vector_valued_regular_least_squares_f_equation}. The Gaussian separable kernel in \eqref{eq:separable_gaussian_kernel} and the odd symplectic kernel in \eqref{eq:odd_symplectic_kernel} were compared for the two systems.

\subsection{Hyperparameter tuning}

The hyperparameters ${\sigma}$ and ${\lambda}$ were tuned for each model by minimizing the cross-validation error \cite{Kohavi1995} over the data set ${\calZ = \{ (\boldx_{i},\boldy_{i}) \in \Rn \times \Rn \}_{i=1}^{N}}$. The data set ${\calZ}$ was split into mutually exclusive subsets ${\calZ_1, \dots \calZ_k}$, and for each iteration ${i \in \{ 1, \dots, k \}}$, the model was trained on the subset ${\hat{\calZ}_{i} = \calZ \setminus \calZ_i}$ and evaluated on ${\calZ_i}$. Formally, the hyperparameter optimization is written as \cite{Krejnik2012}
\begin{equation}
    \min_{\sigma,\lambda} \frac{1}{k} \sum_{i = 1}^{k} \text{MSE}\left( \boldf_{\hat{\calZ}_{i}}, \calZ_i  \right)
\end{equation}
where ${\boldf_{\hat{\calZ}_{i}}}$ is the learned functions trained on the training set ${\hat{\calZ}_{i} = \calZ \setminus \calZ_i}$, using the hyperparameters ${\sigma}$ and ${\lambda}$, ${ \calZ_i}$ is the test set, and MSE is the empirical mean square error between the learned model and the test set. The optimization problem was solved using a grid search.

\begin{figure*}[t!]
    \vspace{4mm}
    \centering
    \begin{subfigure}[h]{0.235\textwidth}
    \centering
        \includegraphics[width=\textwidth]{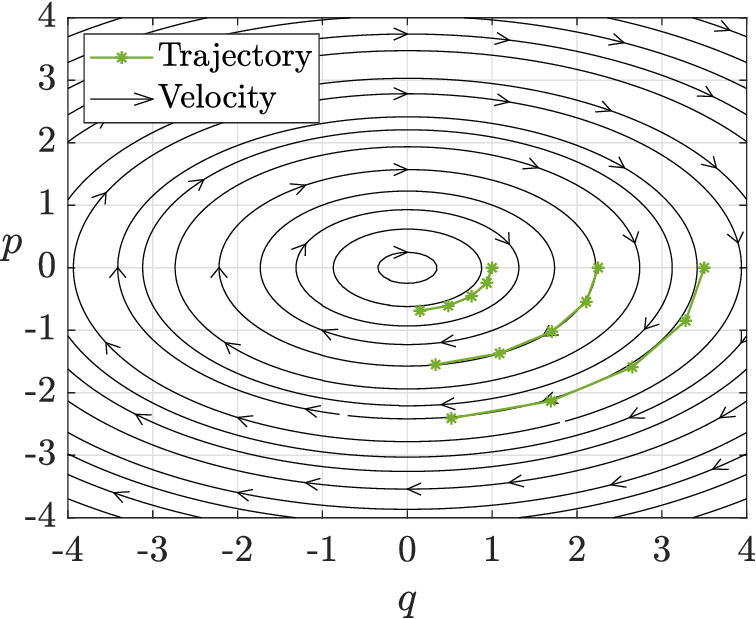}
        \caption{True system}
        \label{fig:harmonic_oscillator_base_model}
    \end{subfigure}
    \hfill
    \begin{subfigure}[h]{0.235\textwidth}
    \centering
        \includegraphics[width=\textwidth]{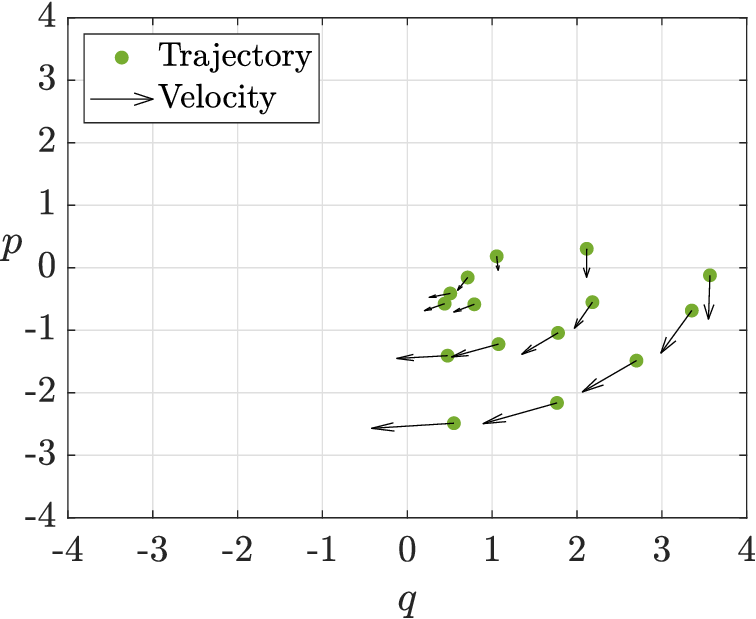}
        \caption{Data set}
        \label{fig:harmonic_oscillator_data_set}
    \end{subfigure}
    \hfill
    \begin{subfigure}[h]{0.235\textwidth}
    \centering
        \includegraphics[width=\textwidth]{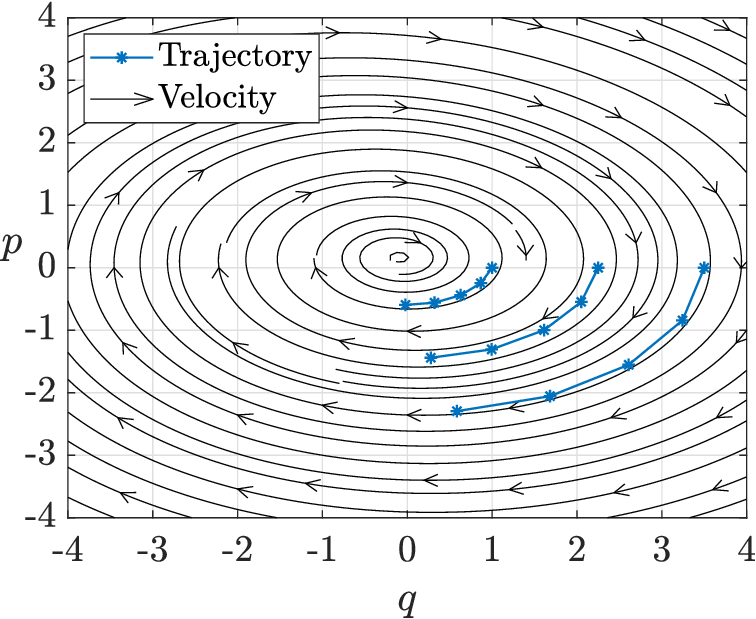}
        \caption{Learned gaussian model}
        \label{fig:harmonic_oscillator_gassuain_kernel}
    \end{subfigure}
    \hfill
    \begin{subfigure}[h]{0.235\textwidth}
    \centering
        \includegraphics[width=\textwidth]{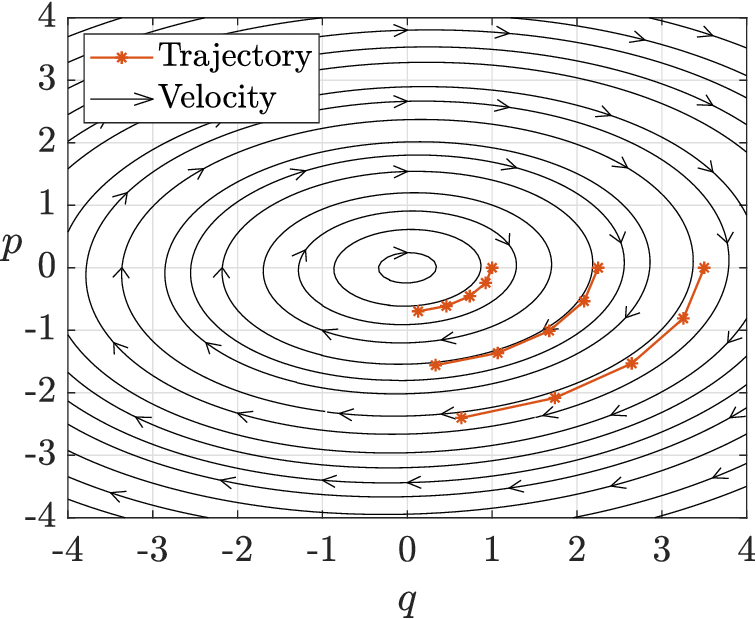}
        \caption{Learned odd sympl. model}
        \label{fig:harmonic_oscillator_odd_symplectic_kernel}
    \end{subfigure}
    \vspace{-1mm}
    \caption{Stream and trajectory plots for the harmonic oscillator and extracted data set, and the resulting learned models using the separable Gaussian kernel and the odd symplectic kernel.}
    \label{fig:harmonic_oscillator}
    \vspace{-3mm}
\end{figure*}

\begin{figure*}[t!]
    \vspace{1mm}
    \centering
    \begin{subfigure}[h]{0.235\textwidth}
    \centering
        \includegraphics[width=\textwidth]{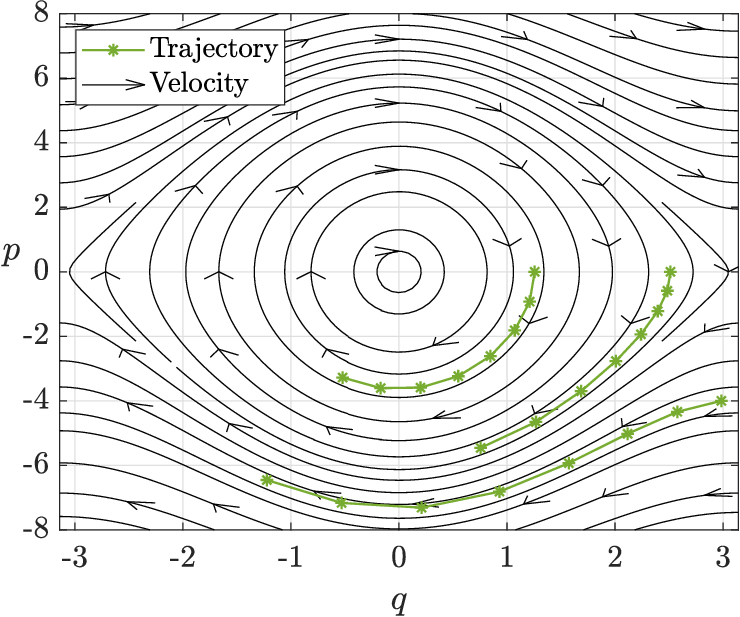}
        \caption{True system}
        \label{fig:simple_pendulum_base_model}
    \end{subfigure}
    \hfill
    \begin{subfigure}[h]{0.235\textwidth}
    \centering
        \includegraphics[width=\textwidth]{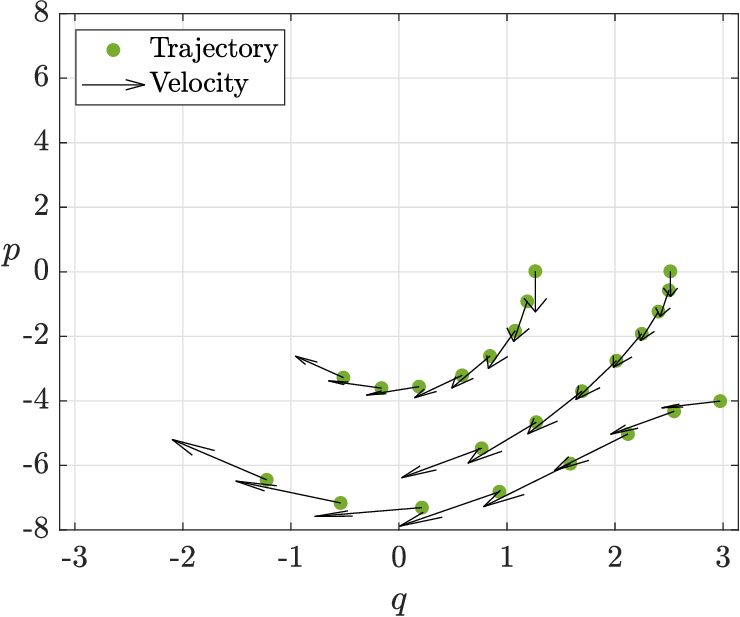}
        \caption{Data set}
        \label{fig:simple_pendulum_data_set}
    \end{subfigure}
    \hfill
    \begin{subfigure}[h]{0.235\textwidth}
    \centering
        \includegraphics[width=\textwidth]{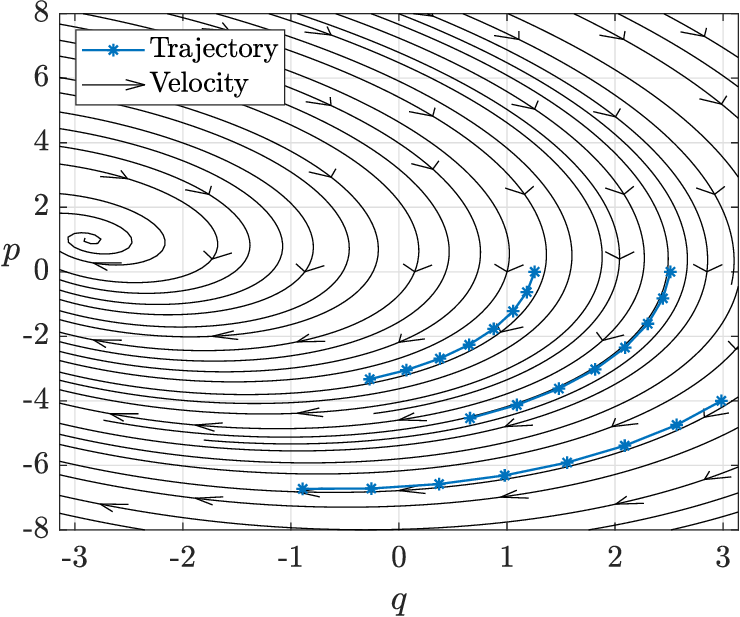}
        \caption{Learned gaussian model}
        \label{fig:simple_pendulum_gassuain_kernel}
    \end{subfigure}
    \hfill
    \begin{subfigure}[h]{0.235\textwidth}
    \centering
        \includegraphics[width=\textwidth]{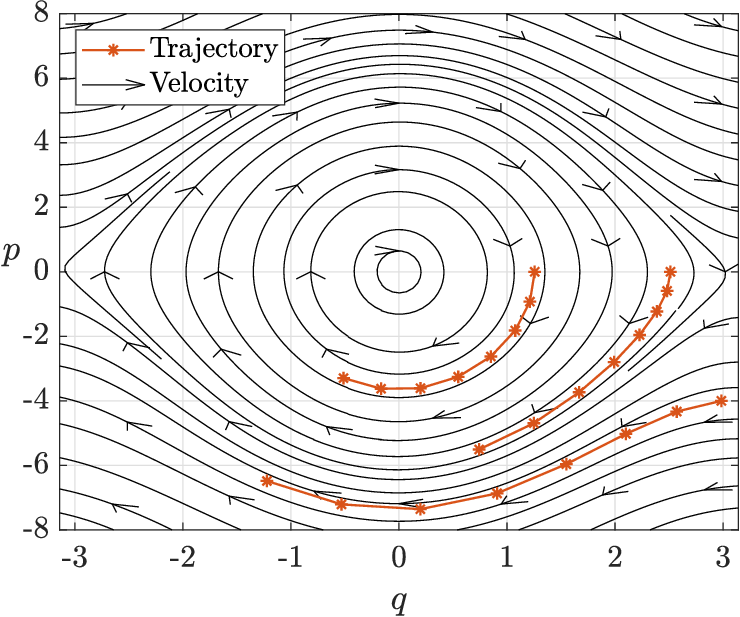}
        \caption{Learned odd sympl. model}
        \label{fig:simple_pendulum_odd_symplectic_kernel}
    \end{subfigure}
    \vspace{-1mm}
    \caption{Stream and trajectory plots for the simple pendulum and extracted data set, and the resulting learned models using the separable Gaussian kernel and the odd symplectic kernel.}
    \label{fig:simple_pendulum}
    \vspace{-5mm}
\end{figure*}

\subsection{Harmonic oscillator}

A harmonic oscillator is given as the undamped mass-spring system 
\begin{equation}
    \xdot = v, \qquad \vdot = - \frac{k}{m} x
\end{equation}
where ${m}$ is the mass, ${k}$ is the spring constant, ${x}$ is the position of the mass, and ${v}$ is the velocity of the mass. The kinetic energy is ${T = \frac{1}{2} m \xdot^2}$ and the potential energy is ${U = \frac{1}{2} k x^2}$. The generalized coordinate is selected as ${q = x}$ and the momentum is then ${p = m \xdot}$. The state vector is $\boldx = [q,p]^\T$. The Hamiltonian for the system is 
\beq\label{eq:harmonic_oscillator_hamiltonian}
    H(q,p) = T(p) + U(q) = \frac{1}{2} \frac{p^2}{m} + \frac{1}{2} k q^2
\eeq
The Hamiltonian dynamics of the true system are then given by
\begin{equation}
    \qdot = \dfrac{\partial H}{\partial p} = \frac{p}{m}, \qquad \pdot = -\dfrac{\partial H}{\partial q} = -k q
\end{equation}
Figure~\ref{fig:harmonic_oscillator_base_model} shows phase curves of the true system as described with parameters ${m = 0.5}$ and ${k = 1}$. Figure~\ref{fig:harmonic_oscillator_base_model} also shows three trajectories that were generated by simulation of the true system with three different initial conditions: ${\boldx_{1,0} = \bb 1, 0\eb^{\T}}$, ${\boldx_{2,0} = \bb 2.25, 0\eb^{\T}}$, and ${\boldx_{3,0} = \bb 3.5, 0\eb^{\T}}$. The time step was ${h = \SI{0.25}{\second}}$ and the system was simulated for ${t \in \bb 0, 1 \eb}$ seconds, which resulted in ${5}$ data points for each trajectory, and ${N = 15}$ total data points. The velocities ${\boldy}$ were sampled at each trajectory point, and zero mean Gaussian noise with standard deviation ${\sigma_n = 0.1}$ was added to the trajectory and velocity data. Figure~\ref{fig:harmonic_oscillator_data_set} shows the resulting data set.

Cross-validation was used for tuning of the hyperparameters, and gave a kernel width ${\sigma_{g} = 19.5}$ and regularization parameter ${\lambda_{g} = 10^{-4}}$ for the Gaussian separable model, and kernel width ${\sigma_{\mathrm{os}} = 12.1}$ and regularization parameter ${\lambda_{\mathrm{os}} = 10^{-4}}$ for the odd symplectic model.

Figures~\ref{fig:harmonic_oscillator_gassuain_kernel} and \ref{fig:harmonic_oscillator_odd_symplectic_kernel} show the learned models when the separable Gaussian kernel and the odd symplectic kernel were used. Both models generalized well, but the odd symplectic model recreated the periodic orbits of the true system shown in Figure~\ref{fig:harmonic_oscillator_base_model}, indicating that the energy conservation was captured in the model. The lack of periodic orbits in Figure~\ref{fig:harmonic_oscillator_gassuain_kernel} documents a lack of energy conservation in the separable Gaussian model.

\begin{figure}[hbt!]
    \centering
    \begin{subfigure}[h]{0.8\columnwidth}
    \centering
        \includegraphics[width=\textwidth]{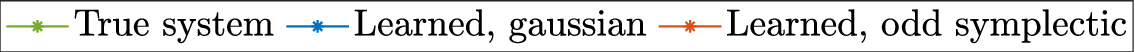}
    \end{subfigure}\\
    \vspace{2mm}
    \begin{subfigure}[h]{0.32\columnwidth}
    \centering
        \includegraphics[width=\textwidth]{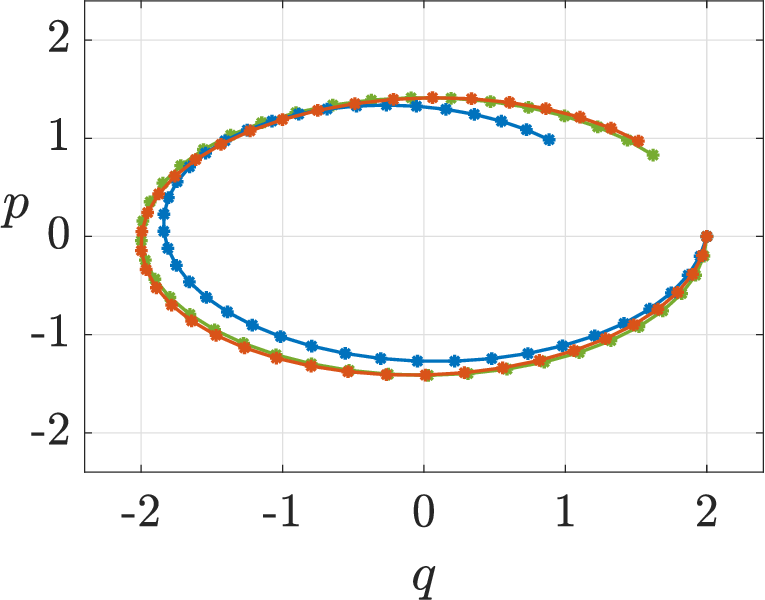}
        \caption{Test trajectory}
        \label{fig:harmonic_oscillator_test_trajectory}
    \end{subfigure}
    \hfill
    \vspace{2mm}
    \begin{subfigure}[h]{0.635\columnwidth}
    \centering
        \includegraphics[width=\textwidth]{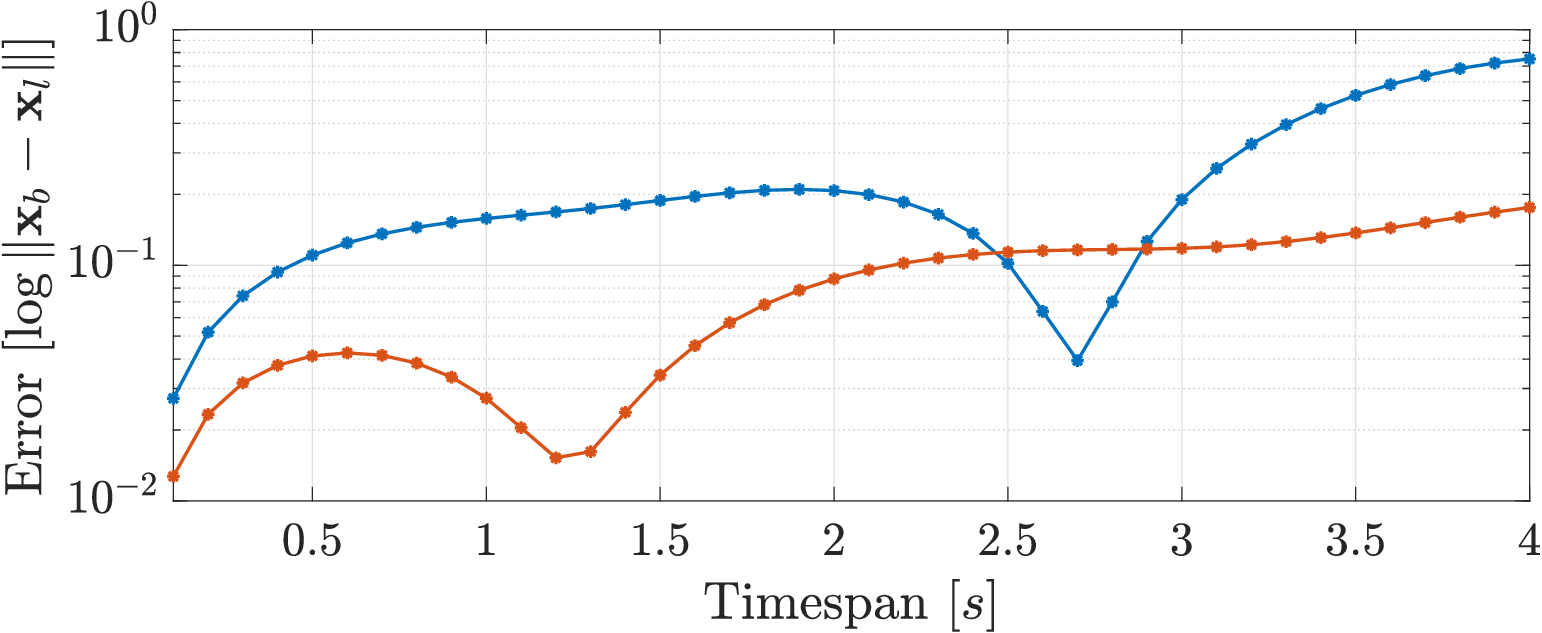}
        \caption{Test trajectory error}
        \label{fig:harmonic_oscillator_test_trajectory_error}
    \end{subfigure}
    \caption{Comparison of the two learned models against the harmonic oscillator system, using the test trajectory.}
    \label{fig:harmonic_oscillator_test}
    \vspace{-6mm}
\end{figure}

A separate test trajectory was simulated to test how well the learned models generalize for unknown data. The initial condition was ${\boldx_{0} = \bb 2, 0\eb^{\T}}$ and the time horizon was ${t \in \bb 0, 4 \eb}$ seconds. The error between the true system and the learned model trajectories is given by ${\text{Err} = \| \boldx_{b} - \boldx_{l} \| }$. Figure~\ref{fig:harmonic_oscillator_test_trajectory} shows the three resulting trajectories, and Figure~\ref{fig:harmonic_oscillator_test_trajectory_error} shows the error for each time step. The results show that the odd symplectic model tracked the true system trajectory with an order of magnitude improvement in accuracy, which indicates the improvement in generalization with the odd, symplectic kernel.

\subsection{Simple pendulum}

A simple pendulum is modeled with a point mass ${m}$ at the end of a mass-less rod of length ${l}$. The pendulum angle is ${\theta}$. The equation of motion is given by
\beq
    \thetaddot = -\frac{g}{l} \sin(\theta)
\eeq
where ${g}$ is the acceleration of gravity. The generalized coordinate is ${q = \theta}$, the kinetic energy is ${T = \frac{1}{2} m l^2 \qdot^2}$ and the potential energy is ${U = m g l (1 - \cos(q))}$. The Lagrangian is defined as ${L = T - U}$ and from  \eqref{eq:general_momenta} the generalized momentum is
\beq
    p = \frac{\partial L}{\partial \qdot} = m l^2 \qdot
\eeq
The Hamiltonian is
\begin{equation}\label{eq:simple_pendulum_hamiltonian}
    H(q,p) = p \qdot - L = \frac{p^2}{2 m l^2} + m g l (1 - \cos(q))
\end{equation}
The Hamiltonian dynamics are then given by
\begin{equation}
    \qdot = \dfrac{\partial H}{\partial p} = \frac{p}{m l^2}, \qquad \pdot = -\dfrac{\partial H}{\partial q} = - m g l \sin(q)
\end{equation}
Figure~\ref{fig:simple_pendulum_base_model} shows the true system with parameters ${m = 0.5}$, ${l = 1}$, and ${g = 9.81}$. Three trajectories were generated, and the system was simulated with three different initial conditions: ${\boldx_{1,0} = \bb \frac{2\pi}{5}, 0\eb^{\T}}$, ${\boldx_{2,0} = \bb \frac{4\pi}{5}, 0\eb^{\T}}$, and ${\boldx_{3,0} = \bb \frac{19\pi}{20}, -4 \eb^{\T}}$. The time step was set to ${h = \SI{0.1}{\second}}$ and the system was simulated for ${t \in \bb 0, 0.7 \eb}$ seconds, giving ${8}$ data points for each trajectory, and ${N = 24}$ total data points. The velocities ${\boldy}$ were sampled at each trajectory point, and zero mean Gaussian noise with standard deviation ${\sigma_n = 0.01}$ was added to the trajectory and velocity data. Figure~\ref{fig:simple_pendulum_data_set} shows the resulting data set.

The cross-validation procedure gave kernel width ${\sigma_{g} = 12.3}$ and regularization parameter ${\lambda_{g} = 0.1}$ for the Gaussian separable model, and kernel width ${\sigma_{\mathrm{os}} = 3}$ and regularization parameter ${\lambda_{\mathrm{os}} = 10^{-4}}$ for the odd symplectic model.

Figures~\ref{fig:simple_pendulum_gassuain_kernel} and \ref{fig:simple_pendulum_odd_symplectic_kernel} show the learned models using the separable Gaussian kernel and the odd symplectic kernel, respectively. The function learned with the Gaussian separable kernel did not give an accurate representation of the true dynamics from such a limited dataset. The model learned with the odd symplectic kernel was accurate and gave a good representation of the vector field of the simple pendulum system. It is seen from Figure~\ref{fig:simple_pendulum_odd_symplectic_kernel} that symmetry and energy conservation lead to periodic orbits. 

\begin{figure}[ht!]
    \vspace{-1mm}
    \centering
    \begin{subfigure}[h]{0.8\columnwidth}
    \centering
        \includegraphics[width=\textwidth]{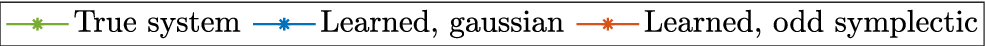}
    \end{subfigure}\\
    \vspace{2mm}
    \begin{subfigure}[h]{0.32\columnwidth}
    \centering
        \includegraphics[width=\textwidth]{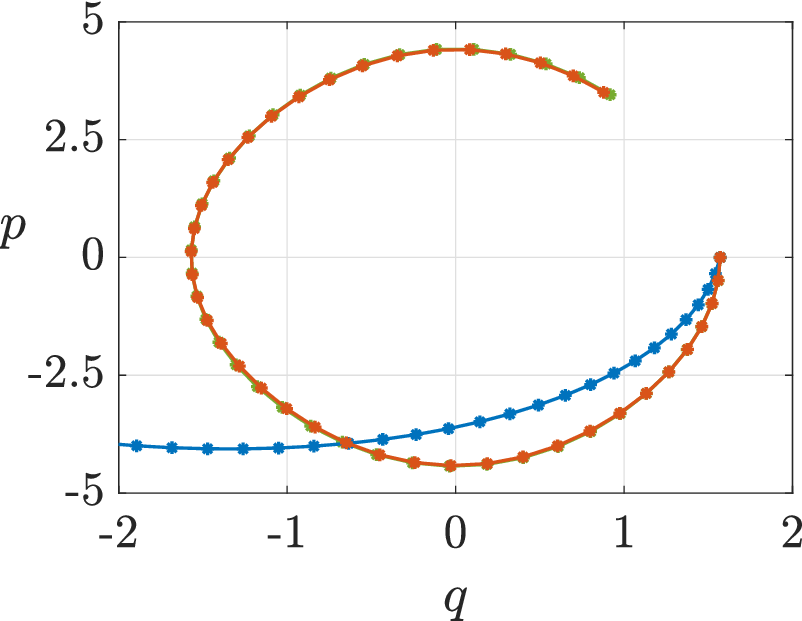}
        \caption{Test trajectory}
        \label{fig:simple_pendulum_test_trajectory}
    \end{subfigure}
    \hfill
    \vspace{2mm}
    \begin{subfigure}[h]{0.635\columnwidth}
    \centering
        \includegraphics[width=\textwidth]{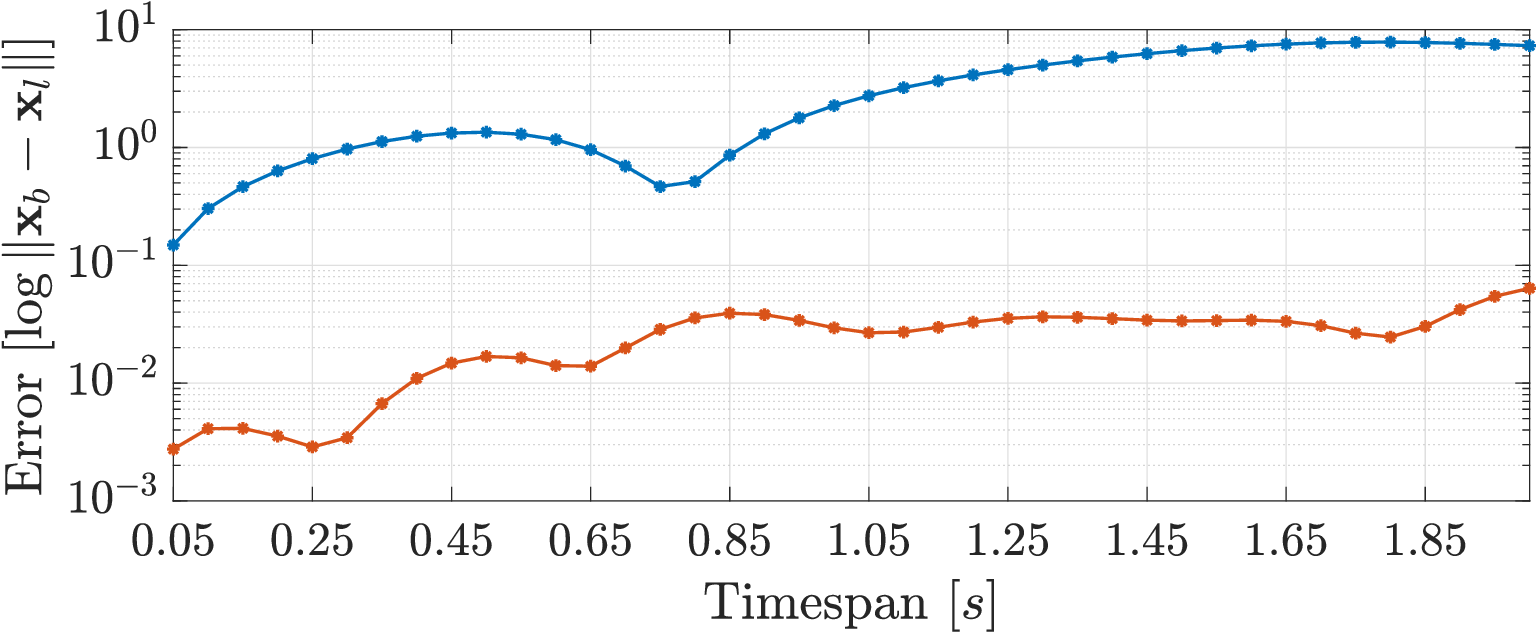}
        \caption{Test trajectory error}
        \label{fig:simple_pendulum_test_trajectory_error}
    \end{subfigure}
    \caption{Comparison of the two learned models against the simple pendulum system, using the test trajectory.}
    \label{fig:simple_pendulum_test}
    \vspace{-5mm}
\end{figure}

A separate test trajectory was simulated to test the generalized performance of the learned models. The initial condition was ${\boldx_{0} = \bb \frac{\pi}{2}, 0\eb^{\T}}$ and the time horizon was  ${t \in \bb 0, 2 \eb}$ seconds. The error between the true system and the learned model trajectories was defined as ${\text{Err} = \| \boldx_{b} - \boldx_{l} \| }$. Figure~\ref{fig:simple_pendulum_test_trajectory} shows the three resulting trajectories, and Figure~\ref{fig:simple_pendulum_test_trajectory_error} shows the error for each time step. The results show that the odd symplectic model is far more accurate than the Gaussian separable model, which fails to generalize beyond the area close to the data set.

\subsection{Numerical evaluation}

The models were evaluated numerically to investigate the ability of the learned models to capture the side information of the true systems. The odd symmetry was evaluated by sampling ${10\,000}$ points in the right half plane of the phase portraits shown in Figure~\ref{fig:harmonic_oscillator} and Figure~\ref{fig:simple_pendulum}, and calculating the odd error given as
\begin{equation}
    e_{\odd} = \| \boldf(\boldx) + \boldf(-\boldx) \|
\end{equation}
where ${\boldf : \Rn \rightarrow \Rn}$ is the dynamical system being evaluated and ${\boldx \in \Rn}$ is the sampled point.

\begin{table}[hbt!]
\vspace{1.34mm}
\caption{Odd error ${e_{\odd}}$ for the two dynamical systems}
\centering
\resizebox{\columnwidth}{!}{
\label{tab:oddness_numerical_evaluation}
\begin{tabular}{@{\extracolsep\fill}lcccccc}
\toprule
& \multicolumn{2}{c}{Harmonic oscillator} & \multicolumn{2}{c}{Simple pendulum}\\
\cmidrule(lr){2-3} \cmidrule(lr){4-5}
System & Mean & Variance & Mean & Variance\\
\midrule
True ${e_{\odd}}$ & 0.00 & 0.00 & 0.00 & 0.00\\
Gaussian sep. ${e_{\odd}}$ & 0.65 & 0.08 & 7.87 & 1.49\\
Odd symplectic ${e_{\odd}}$ & 0.00 & 0.00 & 0.00 & 0.00\\
\bottomrule
\end{tabular}}
\vspace{-5mm}
\end{table}

The results in Table~\ref{tab:oddness_numerical_evaluation} document that the learned odd symplectic model enforces odd symmetry like the true systems, whereas the Gaussian separable model does not. The learned Hamiltonian in \eqref{eq:learned_hamiltonian} for the learned odd symplectic model was evaluated over the test trajectories shown in Figure~\ref{fig:harmonic_oscillator_test} and Figure~\ref{fig:simple_pendulum_test}, and compared to their corresponding real Hamiltonians in \eqref{eq:harmonic_oscillator_hamiltonian} and \eqref{eq:simple_pendulum_hamiltonian}.

The results in Table~\ref{tab:hamiltonian_numerical_evaluation} demonstrate that the value of the learned Hamiltonian ${\hat{H}(\boldx)}$ has a constant offset from the true Hamiltonian ${H(\boldx)}$. This agrees with the fact that the potential energy's zero potential cannot be expected to be the same for the learned and true systems. It is seen that the value of the learned Hamiltonian is constant in agreement with \eqref{eq:hamiltonian_time_derivative_zero} since the system is unforced and independent of time. This is reflected in the variance of both ${H(\boldx)}$ and ${\hat{H}(\boldx)}$. Noting that these are results from numerical simulations, the results indicate that the Hamiltonian mechanics are captured in the learned odd symplectic model.

\begin{table}[hbt!]
\vspace{-2mm}
\caption{Hamiltonian for the two dynamical systems}
\centering
\resizebox{\columnwidth}{!}{
\label{tab:hamiltonian_numerical_evaluation}
\begin{tabular}{@{\extracolsep\fill}lcccc}
\toprule
& \multicolumn{2}{c}{Harmonic oscillator} & \multicolumn{2}{c}{Simple pendulum}\\
\cmidrule(lr){2-3} \cmidrule(lr){4-5}
Hamiltonian & Mean & Variance & Mean & Variance\\
\midrule
Real ${H(\boldx)}$ & $1.99$ & $5.43 \cdot 10^{-9}$ & $9.81$ & $2.2 \cdot 10^{-6}$\\
Learned ${\hat{H}(\boldx)}$ &  $-108.54$ & $6.24 \cdot 10^{-9}$ & $-16.85$& $1.34 \cdot 10^{-6}$\\
\bottomrule
\end{tabular}}
\vspace{-5mm}
\end{table}
\section{CONCLUSION}\label{sec:6_conclusion}

The learning of Hamiltonian mechanical systems with odd vector fields using a specialized kernel has been investigated. The proposed method enforces side information concerning Hamiltonian dynamics and odd vector fields using the kernel rather than through constraints in an optimization problem. The encoding of side information in the kernel allows for a closed-form solution to the learning problem and enforces the constraints over the whole domain of the learned function. Comparative experiments on limited data sets demonstrate how the proposed kernel learns the dynamics of a system compared to a kernel that does not enforce side information. The standard Gaussian kernel recreates the system's dynamics close to the data set, but fails to capture the general characteristics of the system. The proposed kernel enforces side information and produces a model that generalizes well, capturing the dynamics even on a limited and noisy data set.

\textit{Future work}: The proposed method could be applied to more complex mechanical systems. For problems with larger data sets, approximating the kernel using random Fourier features \cite{Rahimi2007} could reduce the problem size, reducing the training and inference time of the learned model. Generalized momenta and their derivatives might not be available in practice, so extending the method to learn Hamiltonian systems from generalized coordinates and velocities only should be investigated. Finally,  control-oriented learning could be studied using the proposed kernel.

\addtolength{\textheight}{-115mm}   





\bibliographystyle{IEEEtran}
\bibliography{utilities/refs}


\end{document}